\documentclass{article}

    \PassOptionsToPackage{numbers, compress}{natbib}

    \usepackage[final]{neurips_2020}

\usepackage[utf8]{inputenc} 
\usepackage[T1]{fontenc}    
\usepackage{hyperref}       
\usepackage{url}            
\usepackage{booktabs}       
\usepackage{amsfonts}       
\usepackage{nicefrac}       
\usepackage{microtype}      
\usepackage{graphicx}
\usepackage{caption}

\title{A More Biologically Plausible Local Learning Rule for ANNs}

\author{%
  Shashi Kant Gupta\\
  Department of Electrical Engineering\\
  IIT Kanpur, UP, 208016 \\
  \texttt{shashikg@iitk.ac.in} \\
}

\begin{document}

\maketitle

\begin{abstract}
  The backpropagation algorithm is often debated for its biological plausibility. However, various learning methods for neural architecture have been proposed in search of more biologically plausible learning. Most of them have tried to solve the "weight transport problem" and try to propagate errors backward in the architecture via some alternative methods. In this work, we investigated a slightly different approach that uses only the local information which captures spike timing information with no propagation of errors. The proposed learning rule is derived from the concepts of spike timing dependant plasticity and neuronal association. A preliminary evaluation done on the binary classification of MNIST and IRIS datasets with two hidden layers shows comparable performance with backpropagation. The model learned using this method also shows a possibility of better adversarial robustness against the FGSM attack compared to the model learned through backpropagation of cross-entropy loss. The local nature of learning gives a possibility of large scale distributed and parallel learning in the network. And finally, the proposed method is a more biologically sound method that can probably help in understanding how biological neurons learn different abstractions.
\end{abstract}

\section{Introduction}
Artificial neural networks have become increasingly popular in the domain of various perceptual tasks, specifically in computer vision \cite{LeCun2015}. Recently, they are also shown to match representations learned in the brain \cite{Yamins2014, Schrimpf407007}. However, its often debated whether the backpropagation algorithm used to learn weights in these networks reflects how the brain learns. Backpropagation has mostly been criticized for 1. its "weight transport problem," i.e., the feedback and forward path needs to be symmetric in weights during learning via backpropagation, which does not seem to be biologically plausible as per current experimental results and 2. "timing problem," i.e., experimental results has shown that the change in biological synapses depends on the spike timing of pre-synaptic and post-synaptic neurons instead of any error between a "fixed target" and "predicted" value \cite{stdp, lillicrap2014random}.

\citet{lillicrap2014random} proposed that random feedback weights support learning in neural networks, and it does not need to be symmetric with the forward connection. This work was further progressed by \citet{dfa}. Although this work solves the weight transport problem, the weight changes still depend on the error between the target value and predicted value, which does not support the STDP learning in biological neurons. In another line of research, \citet{targetprop} proposed that instead of propagating the errors, the network can also learn via target propagation. The target is propagated via a reconstruction layer; therefore weight symmetry problem is solved in this case as well. \citet{sign_symmetric} proposed that the network does not need symmetric weights to learn, but just the sign of weights. The latter two methods also focuses on solving the weight transport problem and did not relate that the change in weights should be dependent on information about spike timing and depends on an error between a target value and predicted value. Biological neurons do not know of any specific target value that it should attain, and magnitude of change in synaptic weights in brain depends on neighboring neurons' spike timing. 

In this work, we investigated a new approach based on the STDP learning rule. Our proposed method covers the salient features of biological learning, which are:
\begin{itemize}
    \item Weight changes depend entirely on the values of pre-synaptic and post-synaptic neurons. We derived our learning rule, assuming that artificial neurons' values model the firing rate of biological neurons.
    \item Whether this weight will add or subtract depends on the neuronal association and will be fixed during the entire training. This is biologically plausible as there's evidence about both hebbian and anti-hebbian type of STDP \cite{antihebb, stdp_review}.
\end{itemize}

Preliminary evaluation on the binary classification task of MNIST \cite{lecun2010mnist} and IRIS \cite{iris} datasets with two hidden layers shows comparable performance with backpropagation. The model learned using this method also shows a possibility of better adversarial robustness against the FGSM attack \cite{fgsm} compared to the model learned through backpropagation of cross-entropy loss.

\begin{figure}[t!]
    \centering
    \includegraphics[width=0.95\linewidth]{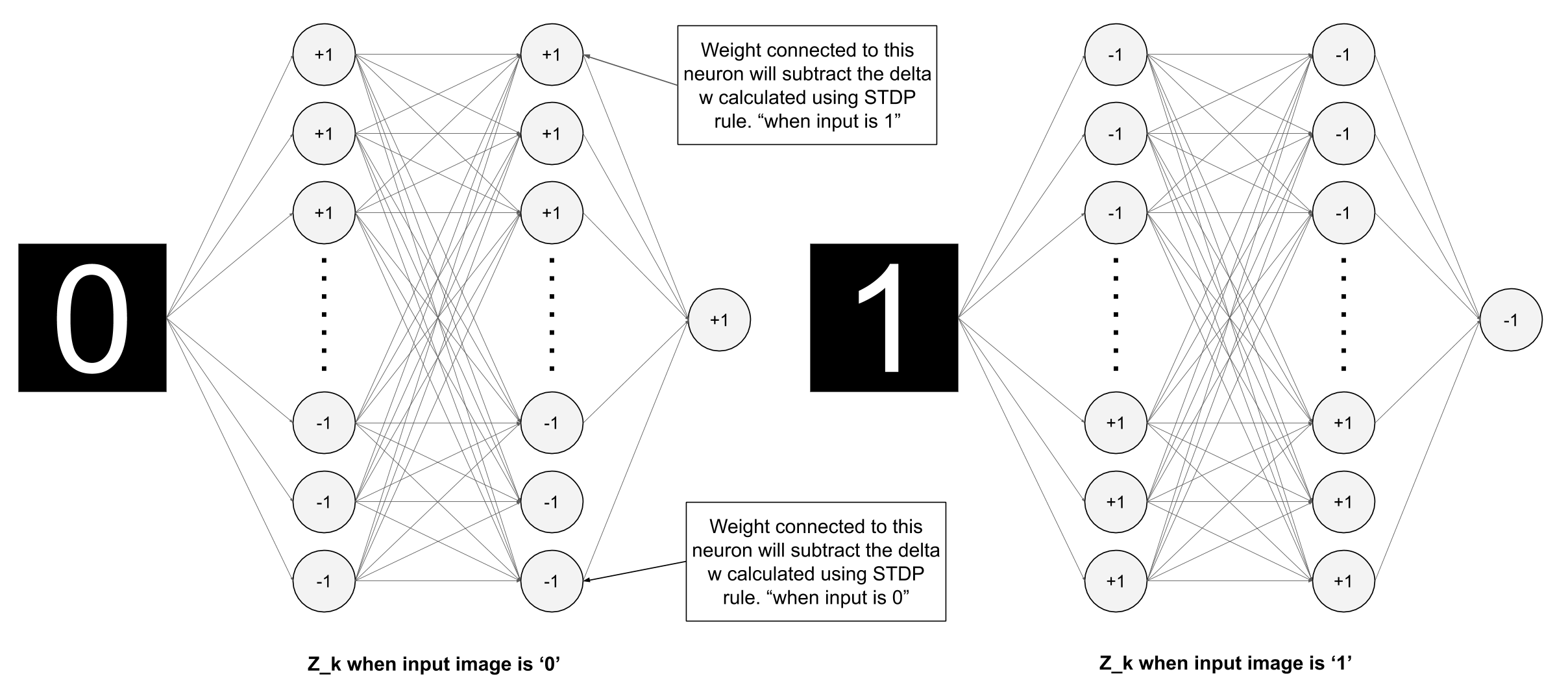}
    \caption{Illustration of how fixed associativity $Z_k$ are assigned in case of a binary classification}
    \label{fig:associativity}
\end{figure}

\section{Spike Timing Dependent Plasticity in ANN}
Spike Timing Dependent Plasticity (STDP) is the most popularly explained model of learning in biological neurons. It says that synaptic weights changes if there are post-synaptic spikes around a pre-synaptic spike. This change is positive if the post-synaptic spike occurs after the pre-synaptic spike and vice-versa. Evidence also suggests the inverse case of this i.e., change is negative if the post-synaptic spike is after the pre-synaptic spike and vice-versa. Also, this change is inversely proportional to the time difference between a pre-synaptic spike and a post-synaptic spike \cite{stdp}.

Since an artificial neuron's output is assumed to be analogous to the firing rate in a biological neuron. Therefore, we can assume spike timing to be inversely proportional to the neuron's output, i.e., $T \propto 1/(firing rate)$. So, based on this we can write $T_{post} \propto 1/p$ and $T_{pre} \propto 1/x$, here $p$ is the value obtained at post-neuron and $x$ is the value obtained at the pre-neuron. Therefore, we can write $\Delta T = T_{post}-T_{pre}$. We know the approximate dependence of weight update as a function of $\Delta T$ in a biological neuron. Based on that we selected the Equation \ref{eq:stdp_eqn} for $\Delta W$. This equation produces the asymmetric version of the STDP curve \cite{Bi1998}. When converted in terms of $x$ and $p$, Equation \ref{eq:stdp_eqn} yields the update rule as shown in Equation \ref{eq:stdp_eqn_xp}. Note that since there's a limitation at what rate a neuron can fire, we use a sigmoid activation instead of a conventional approach to use relu activation (sigmoid will limits the firing rate between 0 and 1). Also, the input data need to be pre-processed, such that its value lies in the range of 0 to 1.

\begin{equation}
    \Delta W \propto \frac{\Delta T}{T_{pre}^2(T_{pre} - \Delta T)^2}
    \label{eq:stdp_eqn}
\end{equation}

\begin{equation}
    \Delta W \propto \frac{x^3p(x - p)}{(2p - x)^2}
    \label{eq:stdp_eqn_xp}
\end{equation}

\textbf{Associativity for supervised learning: } As we can see, the STDP rule results in the weight update rule, which is unsupervised. So, to use it in supervised learning, we use neurons' associativity, which is based on whether the post-neuron wants to associate with the pre-neuron or not to associate. So, our final STDP rule will turn out to be Equation \ref{eq:stdprule}, where $Z_k = +1$ or $Z_k =-1$. We added an $\epsilon = 1e-9$ term in the denominator of Equation \ref{eq:stdprule} to avoid any divisibility by zero term.

\begin{equation}
    \Delta W = +\eta Z_k\frac{x^3p(x - p)}{(2p - x)^2}
    \label{eq:stdprule}
\end{equation}

In a binary classification setting, the $Z_k$ for final output neuron will be $+1$, if $y = 1$ and $Z_k = -1$, if $y = 0$. Where $y$ is the class label. To get the $Z_k$ for hidden layers. The simplest method could be to multiply the next neuron's associativity with the weights between them, and based on the sign, assign them as $-1$ or $+1$. But this will again involve transporting associativity using the synaptic weights between other neurons. And that is against the biological plausibility. What important here to note that when we use feedback connection to get associativity, the network finally on convergence will get a fixed set of $Z_k$, which entirely depends on the initialization of weights. Also, It can be easily observed that if associativity for intermediate layer is $Z_{k}^1$ when input image is of label $y = 1$, then for label $y = 0$ associativity will become $Z_{k}^0 = -Z_{k}^1$. So we can fix this $Z_{k}$ at the beginning of the training and let the network learn its weight to adapt to those associativities. We use one single feedback of whether the output neuron wants to associate or not. This feedback does not depends on weights neither on any specific target value for specific neuron. Therefore, overall this learning method solves most of the issues of biological plausibility. Figure \ref{fig:associativity} shows an illustration of this.

\begin{figure}[t!]
    \centering
    \includegraphics[width=0.95\linewidth]{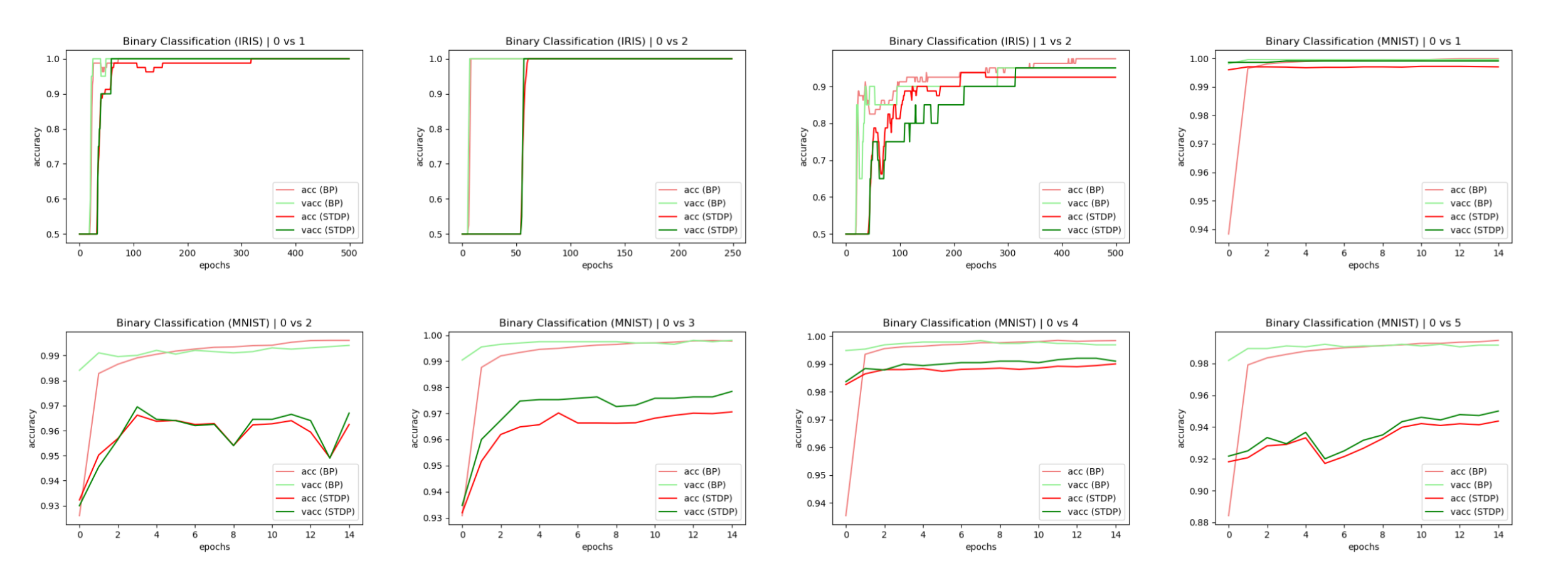}
    \caption{Train and test set accuracy vs epochs for different binary classification pairs. Please note that the y-scale for all the plots shown are different.}
    \label{fig:train_acc}
\end{figure}

\section{Experiment}
We evaluated our method on the binary classification task of MNIST and IRIS datasets. We trained a common architecture for both of them, which contained two hidden layers with the first layer having 32 neurons while the second layer had 16 neurons. We fixed an equivalent number of positive and negative $Z_k$ for each hidden layer. The intuition behind this was that when we trained the network via propagating the association using network weights, we observed that on convergence, this yields an approximately similar number of positive and negative associations. Result for following pairs were shown in the Figure \ref{fig:train_acc}: for IRIS - 0 vs 1, 0 vs 2, and 1 vs 2 and for MNIST - 0 vs 1, 0 vs 2, 0 vs 3, 0 vs 4, and 0 vs 5. We trained the architecture using our (STDP) method and backpropagating binary cross-entropy loss (BP) method. They were trained using adam optimizer with lr=1e-3 with a batch size of 100 in the MNIST case and 40 in the IRIS case. In the STDP case, we replaced the negative gradient with the STDP weight update rule in adam optimizer. We evaluated adversarial accuracy using FGSM attacks at following perturbation strength 4/255, 8/255, and 12/255. We have evaluated the method on other binary pairs as well and the results were similar to these.

\section{Results}
\begin{figure}[t!]
    \centering
    \includegraphics[width=0.99\linewidth]{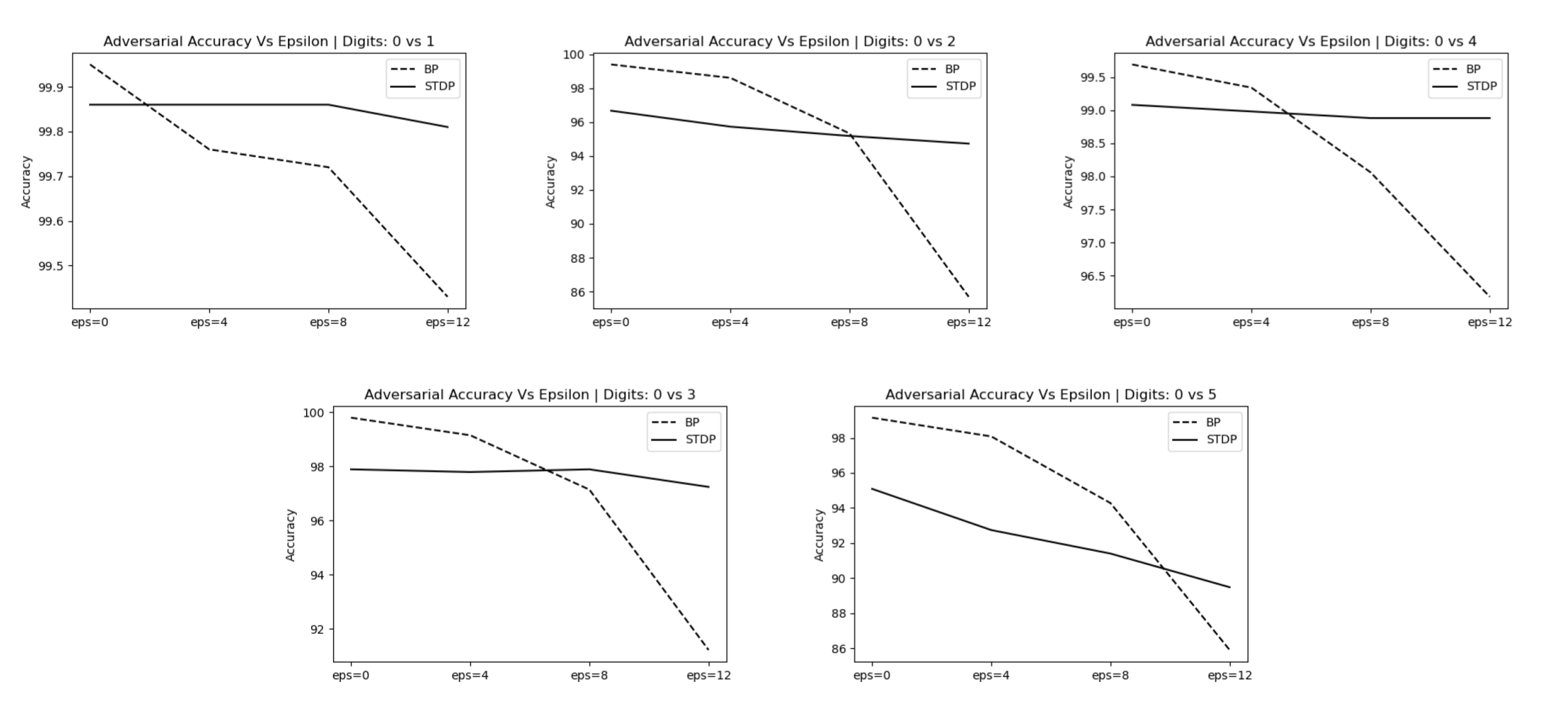}
    \caption{Adversarial accuracy vs perturbation strength for different binary classification pairs. Please note that the y-scale for all the plots shown are different.}
    \label{fig:adv_acc}
\end{figure}

Results from training are shown in Figure \ref{fig:train_acc}. It's quite evident from all the results that the learning method proposed here can learn a multi-layer neural network without any propagation of errors, and with mostly local information of neighboring neurons. The accuracy, when compared to backpropagation, is definitely not equivalent but is quite comparable. For the IRIS dataset where the number of samples is very less, i.e. 120 data points only. The poor performance during initial epochs can be attributed to the fixed nature of associativity. Now looking at the results on adversarial robustness against FGSM attacks (see Figure \ref{fig:adv_acc}), we can see for most cases, this method outperforms the BP method. This advantage in adversarial robustness can be considered to relax the model's slightly poor performance compared to the BP method.

\section{Discussion}
In this work, we proposed a novel method to train neural architecture, independent of any error between the target value and predicted value. We have also shown that it's not needed to propagate any backward signal except information about whether the input stimulus is of the desired class or not. Our proposed method is truly local and incorporates the difference in spikes' timing to estimate its weight updates. The proposed method is probably the next step towards biologically plausible learning in neural networks, which could direct various other research questions in both machine learning and neuroscience community. 

The preliminary evaluation shows a possibility for adversarial robustness. Simultaneously, the local nature of the proposed learning rule can significantly help in distributed and parallel training of neural architecture. Therefore, this method demands further evaluation and testing on various other tasks to test if this could work in different settings.

As a future target, our top-most priority is to adapt this method to the convolution layer. We are not stressing on multi-class classification problems because theoretically if a learning method could learn binary classification problems, it will provide us with a model that could be used for multi-class classification.

\section*{Acknowledgments}
We thank Nisheeth Srivastava for useful discussions and feedback.

\begin{small}
\bibliographystyle{plainnat}
\bibliography{refs}
\end{small}

\newpage
\section*{\LARGE{Appendices}}
\subsection*{A. Intuition Behind The STDP Learning Rule}
As described before we can assume timing of spike to be inversely proportional to output of the neuron i.e. $T \propto 1/(firing rate)$. And based on this we can write $T_{post} \propto 1/p$ and $T_{pre} \propto 1/x$, here $p$ is the value obtained at post-neuron and $x$ is the value obtained at the pre-neuron. Therefore, $\Delta T = T_{post}-T_{pre}$.

Now, for deriving weight update rule we took following two relations: (1) $\Delta W \propto xp$, as per the hebbian postulate we know that the post neuron and pre neuron both associate together and (2) $\Delta W \propto (x - p)$, which is based on the STDP i.e. weight updates should be negative if time difference is negative and vice-versa. So, we get:

\begin{equation}
    \Delta W \propto xp(x - p)
    \label{eq:stdpinit}
\end{equation}

To get further correction in $\Delta W$, we substituted $x$ and $p$ with $1/T_{pre}$ and $1/T_{post}$ in equation \ref{eq:stdpinit} and replaced $T_{post}$ with $T{pre} + \Delta T$. So, as to get the weight updates as function of $\Delta T$ which can be compared with STDP which is found experimentally in biological neurons. This gives us the weight update as shown in equation (\ref{eq:stdpwrong})

\begin{equation}
    \Delta W \propto \frac{\Delta T}{T_{pre}^2(T_{pre} + \Delta T)^2}
    \label{eq:stdpwrong}
\end{equation}

\begin{figure}[h!]
    \centering
    \includegraphics[width=0.99\linewidth]{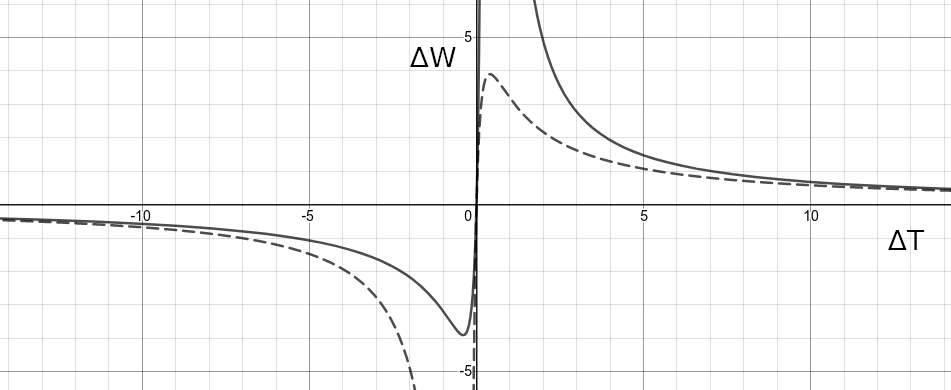}
    \caption{STDP Rule for ANN. Dotted line represents Eqn \ref{eq:stdpwrong} and Solid line represents Eqn \ref{eq:stdpright}}
    \label{fig:stdprule}
\end{figure}

For which when we plotted $\Delta W$ vs $\Delta T$, we got an equivalent graph (Figure \ref{fig:stdprule}, dotted line) as it is found in STDP but the asymmetricity of the graph was opposite to what it is observed in biological neuron i.e. the peak of positive $\Delta T$ should be higher than the peak of negative $\Delta T$. Which is easily visible from the equation (\ref{eq:stdpwrong}) that it is due to positive sign in the denominator term. So, correcting the denominator term we get new update rule as in Equation \ref{eq:stdpright}. This gives us the similar STDP curve as it is in biological neuron (Figure \ref{fig:stdprule}, solid  line). When converted in terms of $x$ and $p$ yields the update rule as in Equation \ref{eq:stdpfinal}. 

\begin{equation}
    \Delta W \propto \frac{\Delta T}{T_{pre}^2(T_{pre} - \Delta T)^2}
    \label{eq:stdpright}
\end{equation}

\begin{equation}
    \Delta W \propto \frac{x^3p(x - p)}{(2p - x)^2}
    \label{eq:stdpfinal}
\end{equation}

\newpage

\subsection*{B. More Results}

\begin{figure*}[h!]
    \parbox{.49\linewidth}{
        \centering
        \includegraphics[width=0.99\linewidth]{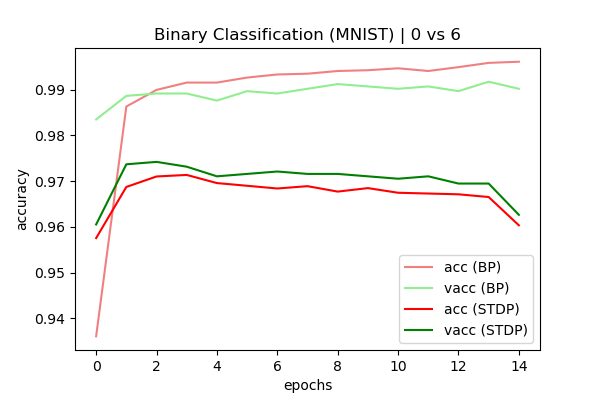}
    }
    \parbox{.49\linewidth}{
        \centering
        \includegraphics[width=0.99\linewidth]{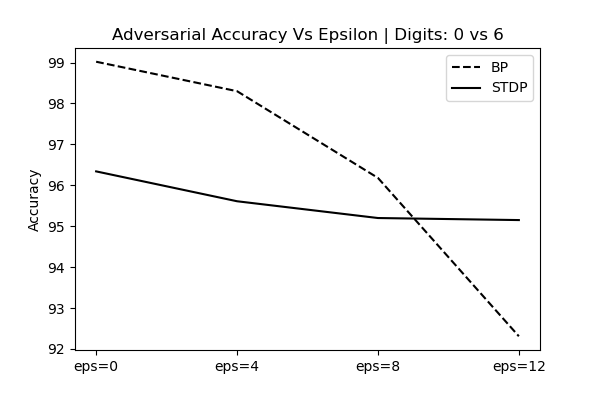}
    }
\end{figure*}

\begin{figure*}[h!]
    \parbox{.49\linewidth}{
        \centering
        \includegraphics[width=0.99\linewidth]{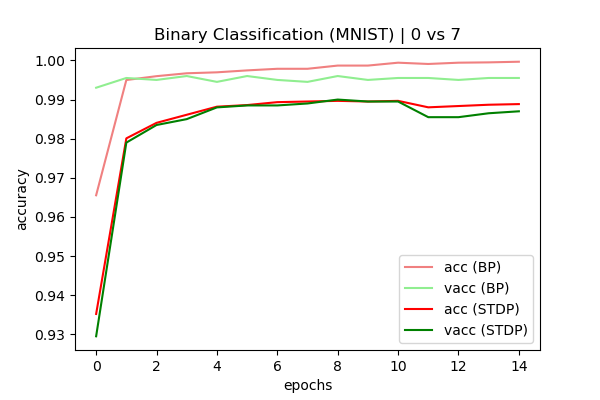}
    }
    \parbox{.49\linewidth}{
        \centering
        \includegraphics[width=0.99\linewidth]{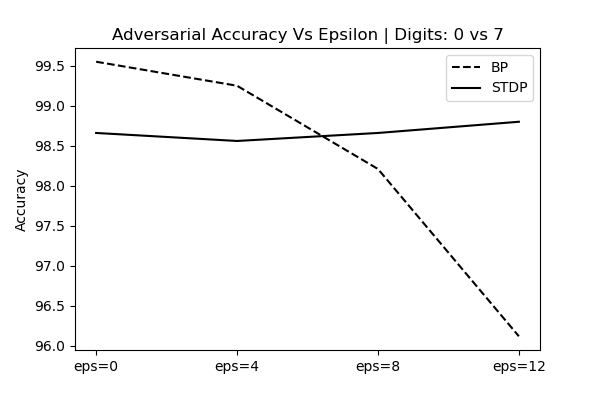}
    }
\end{figure*}

\begin{figure*}[h!]
    \parbox{.49\linewidth}{
        \centering
        \includegraphics[width=0.99\linewidth]{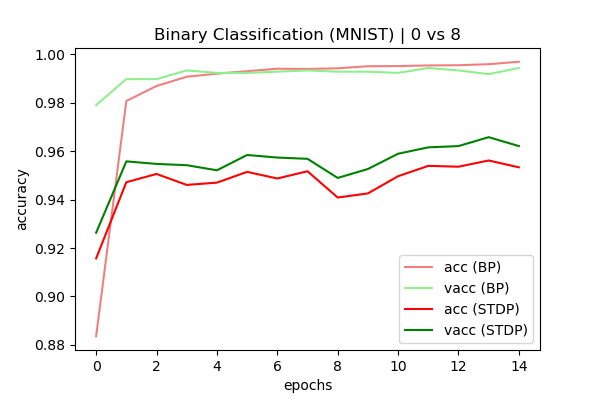}
    }
    \parbox{.49\linewidth}{
        \centering
        \includegraphics[width=0.99\linewidth]{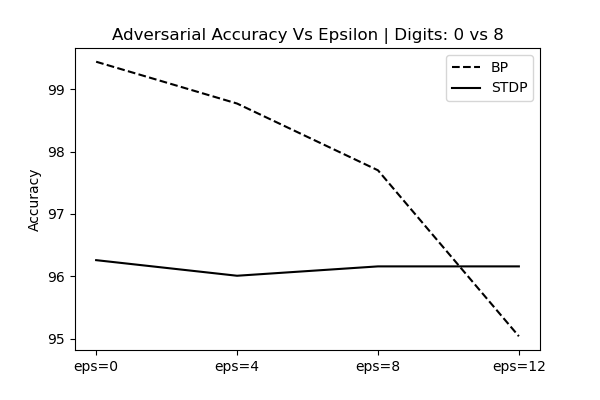}
    }
\end{figure*}

\begin{figure*}[h!]
    \parbox{.49\linewidth}{
        \centering
        \includegraphics[width=0.99\linewidth]{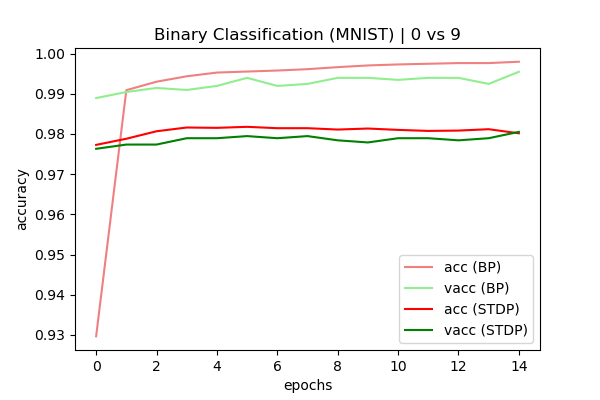}
    }
    \parbox{.49\linewidth}{
        \centering
        \includegraphics[width=0.99\linewidth]{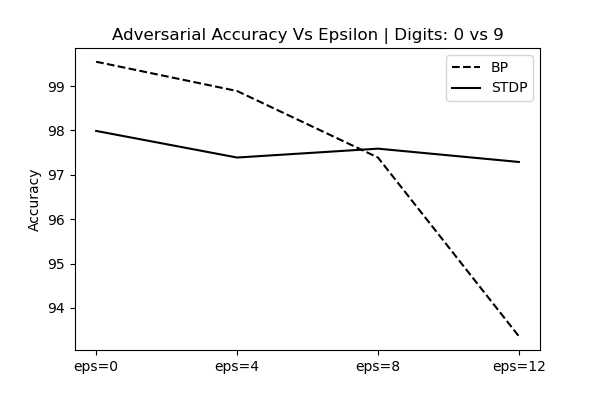}
    }
\end{figure*}

\begin{figure*}[h!]
    \parbox{.49\linewidth}{
        \centering
        \includegraphics[width=0.99\linewidth]{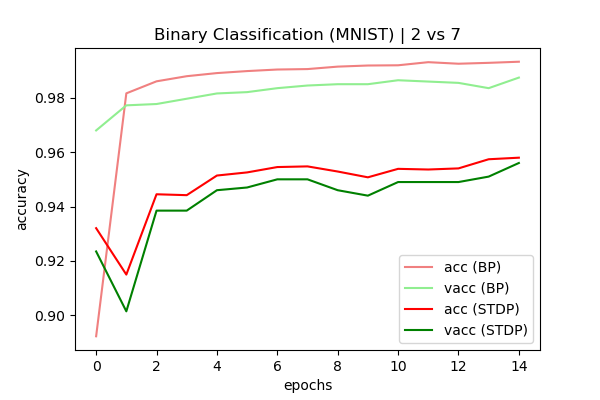}
    }
    \parbox{.49\linewidth}{
        \centering
        \includegraphics[width=0.99\linewidth]{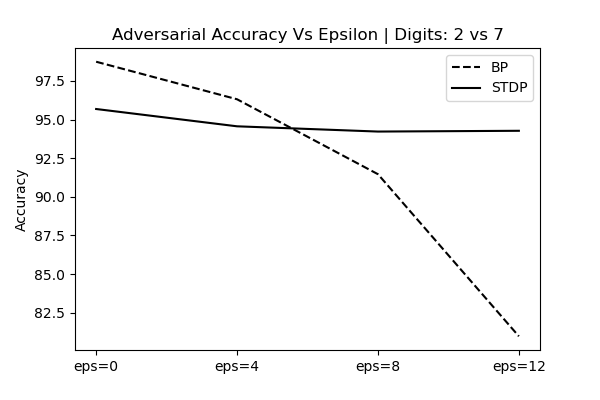}
    }
\end{figure*}

\begin{figure*}[h!]
    \parbox{.49\linewidth}{
        \centering
        \includegraphics[width=0.99\linewidth]{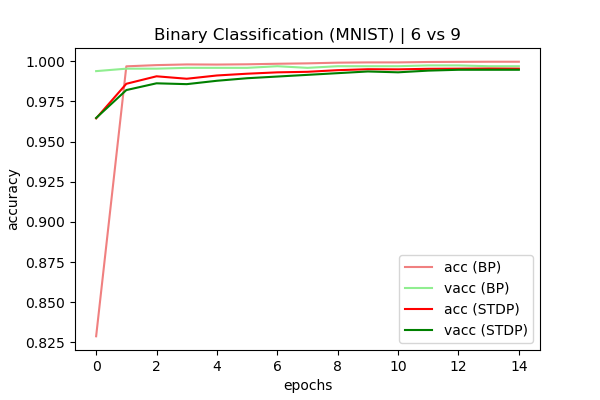}
    }
    \parbox{.49\linewidth}{
        \centering
        \includegraphics[width=0.99\linewidth]{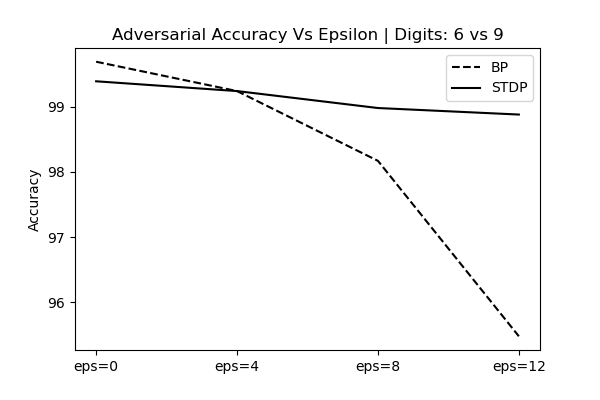}
    }
\end{figure*}

\begin{figure*}[h!]
    \parbox{.49\linewidth}{
        \centering
        \includegraphics[width=0.99\linewidth]{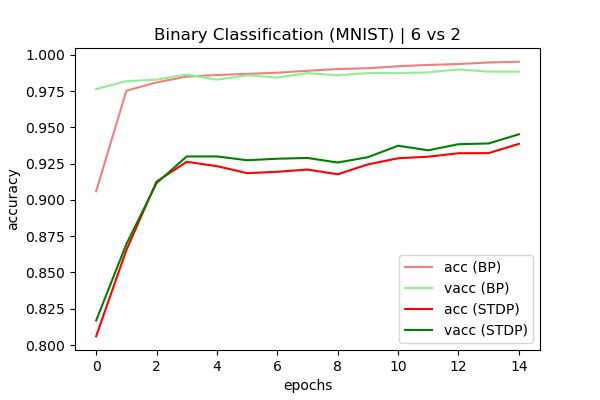}
    }
    \parbox{.49\linewidth}{
        \centering
        \includegraphics[width=0.99\linewidth]{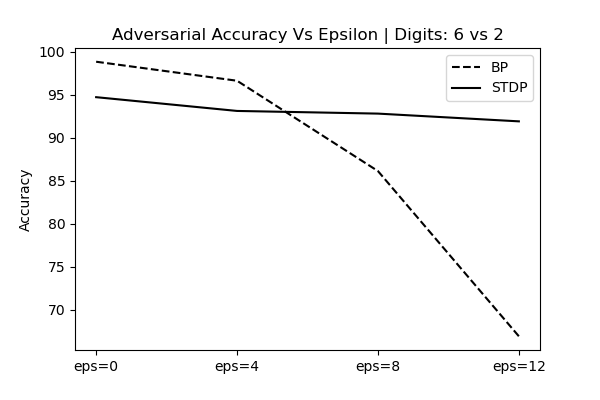}
    }
\end{figure*}

\begin{figure*}[h!]
    \parbox{.49\linewidth}{
        \centering
        \includegraphics[width=0.99\linewidth]{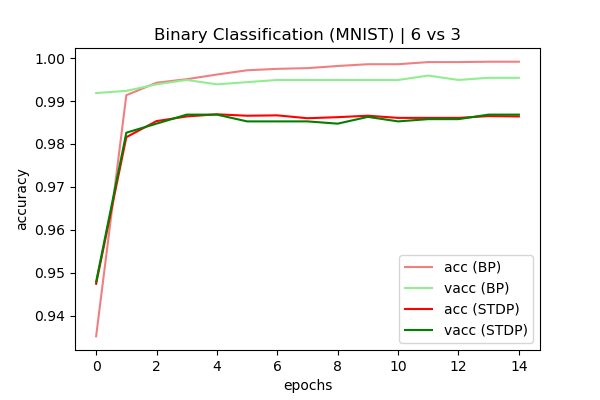}
    }
    \parbox{.49\linewidth}{
        \centering
        \includegraphics[width=0.99\linewidth]{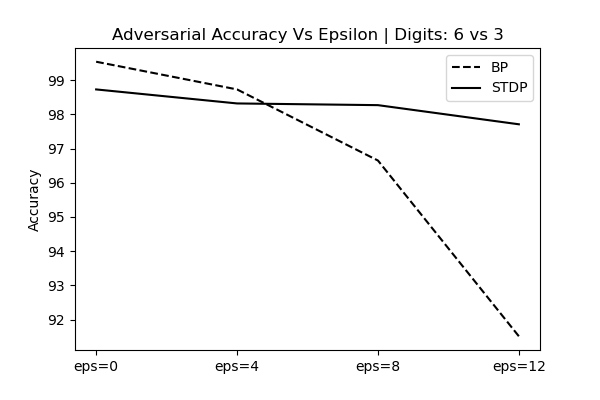}
    }
\end{figure*}

\end{document}